%% file: main.tex
\documentclass[runningheads]{llncs}

\input{bold}

\def\etal{~\textit{et.al.}}

\usepackage{xcolor}
\usepackage{amssymb}
\usepackage{booktabs}
\usepackage{graphicx}
\usepackage{amsmath}
\usepackage{upgreek}
\usepackage{soul}
\usepackage{mathtools}

\usepackage{floatrow}
\usepackage{booktabs}

\DeclareMathOperator*{\argmin}{arg\,min}
\usepackage{breqn}

\usepackage{easyReview}

\begin{document}
%


\title{\textit{OLVA}: \textit{O}ptimal \textit{L}atent \textit{V}ector \textit{A}lignment for Unsupervised Domain Adaptation in Medical Image Segmentation\thanks{This work has been supported in part by the European Regional Development. Fund, the Pays de la Loire region on the Connect Talent scheme (MILCOM Project) and Nantes Métropole (Convention 2017-10470),}}

\author{Dawood Chanti \and Diana Mateus}

\titlerunning{OLVA}
%

\author{Dawood Al Chanti \and Diana Mateus }

\institute{École Centrale de Nantes, Laboratoire des Sciences du Numérique de Nantes LS2N, UMR CNRS 6004 Nantes, France.}



%
\maketitle              
\begin{abstract}

This paper addresses the domain shift problem for segmentation. As a solution, we propose OLVA, a novel and lightweight unsupervised domain adaptation method based on a Variational Auto-Encoder (VAE) and optimal transport (OT) theory. Thanks to the VAE, our model learns a shared cross-domain latent space that follows a normal distribution, which reduces the domain shift. To guarantee valid segmentations, our shared latent space is designed to model the shape rather than the intensity variations. We further rely on an OT loss to match and align the remaining discrepancy between the two domains in the latent space. We demonstrate OLVA's effectiveness for the segmentation of multiple cardiac structures on the public Multi-Modality Whole Heart Segmentation (MM-WHS) dataset, where the source domain consists of annotated 3D MR images and the unlabelled target domain of 3D CTs. Our results show remarkable improvements with an additional margin of $12.5\%$ dice score over concurrent generative training approaches.

\end{abstract}

\section{Introduction}

\begin{figure}[t]
\includegraphics[width=\textwidth]{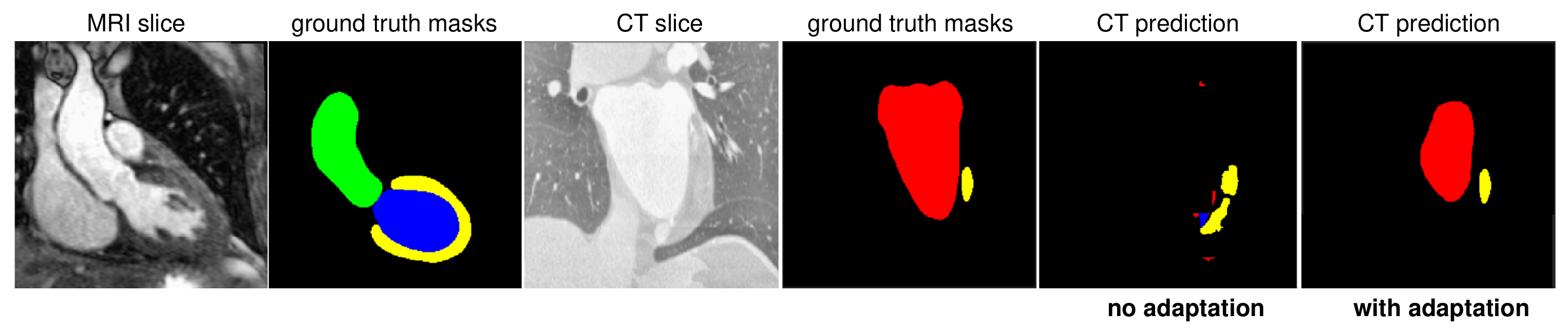}
\caption{The appearances of the cardiac structures look significantly different on MR and CT images. Both modalities share the same label space, yellow: left ventricle myocardium (LV-M), blue: left ventricle blood cavity (LV-B), red: left atrium blood cavity (LA-B), and green: ascending aorta (A-A). Bad prediction due to severe domain shift when no adaptation is considered. MM-WHS cardiac public database \cite{zhuang2016multi}.} \label{fig1}
\end{figure}

Automatic segmentation from multi-modal images is essential for clinical assessment, diagnosis and treatment planning \cite{ackaouy2020unsupervised,ouyang2019data}. Extensive literature has shown the effectiveness of convolutional neural networks in segmenting accurately cardiac structures \cite{li2021mdfa,painchaud2020cardiac}. Yet, without proper adaptation these models fail when deployed across modalities, new subjects and different clinical sites, due to a domain shift \cite{heimann2013learning} \textit{e.g.} between the modalities' appearance as in Fig.~\ref{fig1}. Designing models that can perform well across domains is key in medical applications where labels are scarce and expensive to obtain.

Semi-supervised and Unsupervised Domain Adaptation (UDA) approaches have been proposed to tackle the domain shift problem. The former assume a few labeled instances in the target domain can be used for joint-training with the source data \cite{puybareau2018left}. The more ambitious UDA strategies \cite{ackaouy2020unsupervised,chen2019synergistic,chen2020unsupervised,ouyang2019data,wu2020cf,yang2019domain} assume no labels are available for the target domain. The core idea of UDA is to go through an adaption phase using a non-linear mapping to find a common domain-invariant representation or a latent space $\mathcal{Z}$. The 
domain shift in $\mathcal{Z}$ can be reduced by enforcing the two domains distributions to be closer via a certain 
loss
(\textit{e.g.} Maximum Mean Discrepancy \cite{kumagai2019unsupervised}). Since $\mathcal{Z}$ is common to all domains who share the same label space, projected labeled source domain samples can be used to train a segmenter for all domains. In this paper, we deal with the problem of UDA for MR-CT cross-modality cardiac structure segmentation.

\textbf{Related Work.} Recent works on UDA for medical image segmentation rely on Generative Adversarial Networks  \cite{chen2019synergistic,chen2020unsupervised,dou2018pnp,ouyang2019data,wu2020cf,yang2019domain} to translate the appearance from one modality to the other using multiple discriminators and a pixel-wise cycle consistency loss. Despite their success, they: i) suffer from instabilities \cite{arjovsky2017towards}, ii) rely on complex architectures with more than 95 million parameters, iii) are prone to model collapse \cite{liu2019spectral}, and iv) may generate images outside the actual target domain \cite{ackaouy2020unsupervised}. To alleviate some of these limitations, Ouyang\etal~\cite{ouyang2019data} combined adversarial networks with VAEs~\cite{kingma2013auto}. They exploited the VAEs constraint imposed on the latent space to match a prior distribution and experimentally validated that it reduces the domain shift when used as a shared space across domains. To encourage appearance-invariance, \cite{ouyang2019data} deployed an adversarial loss guided by a cycle-consistency. The VAE model in \cite{ouyang2019data} is complex as equipped with three encoders, three decoders, a segmenter, and a domain classier. Its loss function has six trade-off hyper-parameters to tune. Recently, Optimal Transport (OT) theory \cite{villani2008optimal} jointly with deep learning methods was used by Ackaouy\etal~\cite{ackaouy2020unsupervised} where a joint cost measure combining both the distances at feature space of a deep 3D-Unet between the samples and a loss function measuring the discrepancy at the output space between the two domains is proposed. A limitation of Seg-JDOT is that it employs image patches to enable the generation of a higher number of samples and to avoid curse of dimensionality when optimizing for the transport plan $\gamma$. 

\textbf{Proposal.} We present a novel and lightweight domain-invariant variational -segmentation auto-encoder model. We use the latent space of a VAE that is constrained to follow a prior normal distribution as a common space similar to \cite{ouyang2019data} to reduce the domain shift. Then we exploit the geometry in $\mathcal{Z}$ for matching and aligning distances between probability distribution using OT theory by optimizing for a transport plan $\gamma$, similar to \cite{ackaouy2020unsupervised}, to further shrink the remaining domain shift. Different from \cite{ouyang2019data}, who maximized the image likelihood, we directly learn a semantic latent representation that maximizes the label likelihood. Our idea is that the prior normal distribution has a limited capacity to handle intensity and shape variations, but it can be efficiently exploited for modelling shapes alone. This claim is supported by \cite{painchaud2020cardiac} who use a VAE as a post-processor on the top of a U-Net output to convert the erroneous U-Net predictions to anatomical plausible outputs. Conversely, we simultaneously perform anatomical plausible segmentation and partial alignment of the label-conditional distributions. Also, different from \cite{ackaouy2020unsupervised}, i) we operate over the full image scale, ii) we bring the source and the target data closer to a normal distribution before solving for $\gamma$ to guarantee its convergence and iii) we do not require the alignment of the label-conditional distributions at the label space.

\textbf{\textit{Our main contributions are}:} i) We reduce the domain shift between the source and the target domain by projecting them into a shared semantic latent space which is regularized to follow a prior normal distribution; ii) We address the remaining shift by aligning latent vectors from both domains using a discrepancy measure based on OT theory; iii) Different from a typical VAE which forces the latent space to model image intensity variations, we concentrate the limited capacity of the prior normal distribution to model the shape of the segmentation masks; iv) Our model is lightweight with 1.7 million parameters and easy to adapt for other clinical application; v) We validate our model on the MM-WHS public dataset and outperform state of the art methods by a margin of $12.5\%$ dice score.
\section{Method}

\begin{figure}[t]
\includegraphics[width=\textwidth]{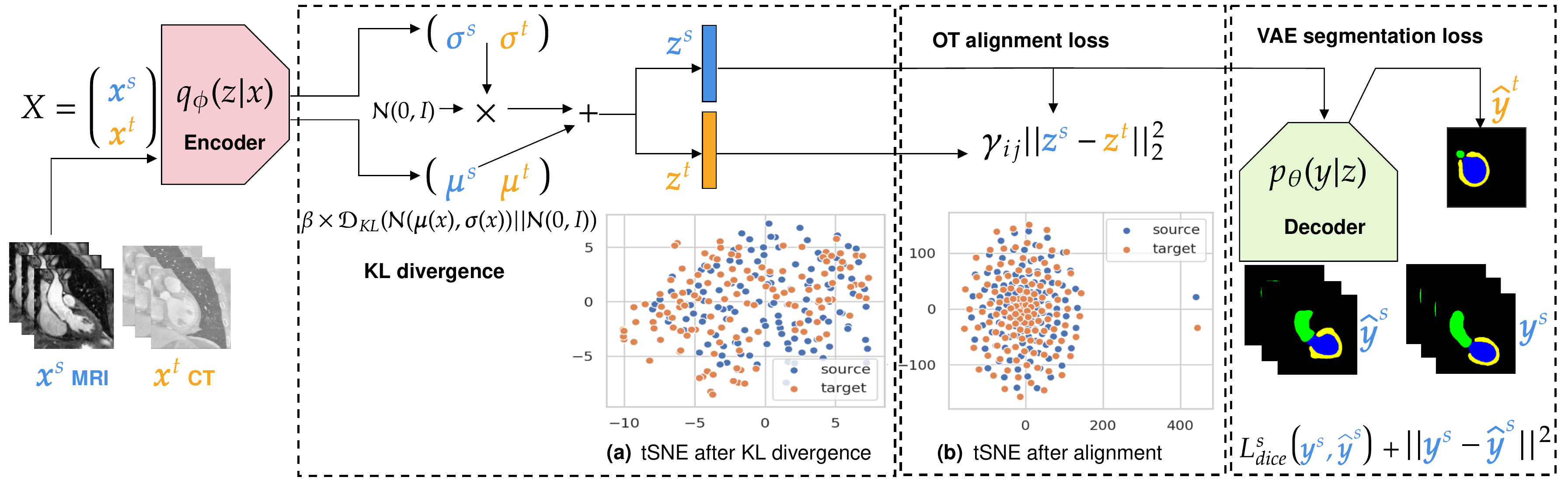}
\caption{OLVA: a generative encoder enforces the shared latent vectors of both source and target domain to follow a prior normal distribution (a), further aligned through an OT plan $\gamma$ (b). A generative decoder is guided with the source domain labels to produce segmentation maps. We use t-distributed stochastic neighbor embedding (t-SNE) 
to map the latent vectors to a 
2D space for visualization purposes only.} \label{fig2}
\end{figure}

Consider a labeled source domain dataset $\{{X}^{s},{Y}^{s}\}_{s=1}^N$ with $N$ images, and a target dataset $\{{X^t}\}_{t=1}^M$ with $M$ images, but with unknown labels ${Y}^{t}$. The goal of UDA is to build a common space for ${X}^{s},{X}^{t}$ while using the source labels ${Y}^{s}$ to guide a segmentation model to generalize across both domains. Here, we propose an Optimal Latent Vector Alignment (OLVA) method to learn a shared latent space that encodes all the structural information needed to generate image segmentation masks, regardless of the domain. We minimize the domain shift between the source and the target distributions in this latent space by pushing them close to a prior normal distribution with a VAE. An optimal transport discrepancy measure removes the remaining domain shift. Finally, a generative decoder guided by the source labels is trained to produce feasible semantic segmentation masks. A  block diagram of the method is shown in Fig.~\ref{fig2}.

\subsection{VAEs for segmentation}

The goal of VAEs is to search for the best parameters $\phi^\ast,\theta^\ast$ in order to sample a latent variable $\boldsymbol{z}\sim q_{\phi^\ast}(\boldsymbol{z|x})$ whose
distribution can be relatively simple such as isotropic Gaussian distribution and to generate a new sample $\boldsymbol{\hat x}\sim p_{\theta^\ast}(\boldsymbol{z|x})$ as close as possible to the real observed data $\boldsymbol{x}$ such that $p_\theta^\ast(\boldsymbol{x|z})=p^\ast(x)$. The VAE loss formalized as in Eq.~(\ref{eq1}) enables an end-to-end training with a first term that maximize the marginal likelihood so that the generative model becomes better and a regularization term that minimize KL-divergence to better approximate $q_{\phi}(\boldsymbol{z|x})$ from the posterior $p_{\theta}(\boldsymbol{z})$. $\beta$ is a trade-off parameter between the two terms. Commonly, a prior model $p(\boldsymbol{z})$ is set to normal distribution $\mathcal{N}(0;I)$ and the re-parametrization trick is applied to facilitate the sampling process as $q_{\phi}(\boldsymbol{z|x})=\mathcal{N}(\mu_{\phi}(\boldsymbol{x}),\sigma_{\phi}(\boldsymbol{x})I)$. Thereby, $\mathcal{D}_{KL}$ become equivalent to $\frac{1}{2} \sum_{k=1}^{K} \left(1+\log(\sigma^{2}_{k}) - \mu^{2}_{k}-  \sigma^{2}_{k}\right)$, $K$ is the latent space dimension and $-\log p_{\theta}(\boldsymbol{x}|\boldsymbol{z})$ is conveniently replaced by a reconstruction loss $||\boldsymbol{x}- p_{\theta}(\boldsymbol{x}|\boldsymbol{z})||^{2}$.



\begin{equation}
\mathcal{L}_{vae}(\phi,\theta;\boldsymbol{x}) =\underset{z \sim q_{\phi}}{\mathbb{E}} \left[-\log p_{\theta}(\boldsymbol{x|z}) \right] + \beta\mathcal{D}_{KL}(q_{\phi}(\boldsymbol{z|x})||p_{\theta}(\boldsymbol{z})), 
\label{eq1}
\end{equation}



To use VAEs for segmentation, we let $\boldsymbol{z}$ represent directly the latent space of the segmentation masks, since modelling shape alone is less complex than shape and intensity together.
Moreover, VAEs lead to latent spaces that are continuous and structured (and not discrete as those of U-Net-like networks), facilitating interpolation. These choices will be determinant in producing valid segmentation masks that respect the anatomical variations of the source domain. 
In practice, our decoder acts as a predictive generative model for the conditional label distribution $p_{\theta}(\boldsymbol{y|z})$ defined on the label space $\mathcal{Y}$. Using source domain data $\{X^s,Y^s\}$, our VAE segmentation loss is also guided by the soft dice loss $\mathcal{L}_{dice}^{s}$ to provide predictions of the segmentation maps as shown in Eq.~(\ref{eq2}.)

\begin{equation}
\mathcal{L}_{vae}^{s} = ||\boldsymbol{y}^{s} - p(\boldsymbol{y}^{s}|\boldsymbol{z}^{s})||^{2} +\beta\mathcal{D}_{KL}^{s}(q(\boldsymbol{z}^{s}|\boldsymbol{x}^{s})||p_{\theta}(\boldsymbol{z}^{s}))+ \mathcal{L}_{dice}^{s}(\boldsymbol{y}^{s},p(\boldsymbol{y}^{s}|\boldsymbol{z^{s}})).
\label{eq2}
\end{equation}


\subsection{Optimal Transport for Latent Vector Alignment}

To solve the domain adaptation problem within the segmentation task, we assume the existence of two distinct joint probability distributions $\mathcal{P}^s$ and $\mathcal{P}^t$ defined over a shared latent space $\mathcal{Z}$ and their marginal distributions ($\zeta^{s},\zeta^{t}$) are defined over $\Omega$ (the set of all probability measures). 
Using $\mathcal{L}_{vae}^{s}(\phi,\theta;\boldsymbol{x}^{s},\boldsymbol{y}^{s})$, we regularize $\mathcal{P}^s$ to follow a normal distribution. To partially align $\mathcal{P}^s$ and $\mathcal{P}^t$, we argue that forcing both distribution to the same prior is beneficial. Therefore, we impose an additional cost function $\mathcal{D}_{KL}^{t}$ that measures the dissimilarity between latent vectors from the source $q(\boldsymbol{z}^s|\boldsymbol{x}^{s})$ and the target $q(\boldsymbol{z}^t|\boldsymbol{x}^{t})$ domains, with $\boldsymbol{z}^s, \boldsymbol{z}^t \in \mathcal{Z}$. Although necessary, this step is insufficient to completely align the two domains. The optimal transport (or Monge-Kantorovich) \cite{ackaouy2020unsupervised,OTCourty,gopalan2011domain,villani2008optimal} problem involves the matching of probability distributions defined over a geometric domain such as our latent semantic space. Here, using OT theory, we seek for a transportation plan matching distributions $\mathcal{P}^s$ and $\mathcal{P}^t$, which is equivalent to finding a probabilistic coupling, $\gamma_{0} \in \prod(\mathcal{P}^s,\mathcal{P}^t)$, as shown in Eq.~(\ref{eq3}). To simultaneously align the latent space through a coupling $\gamma_{0}$ while optimizing for $q(\boldsymbol{z|x})$, we adapt the Kantorovich OT formulation to the discrete case as in Eq.~(\ref{eq4}),
%
%
%

\begin{align}
    \gamma_{0} = \argmin_{\prod(\mathcal{P}^s,\mathcal{P}^t)} \int_{\Omega \times \Omega} \mathcal{D}(q(\boldsymbol{z^s}|\boldsymbol{x^{s}});q(\boldsymbol{z^t}|\boldsymbol{x^{t}}))d\gamma(q(\boldsymbol{z^s}|\boldsymbol{x^{s}});q(\boldsymbol{z^t}|\boldsymbol{x^{t}})). \label{eq3}
\\  
\min_{\gamma \in \prod, q(\boldsymbol{z}|\boldsymbol{x})
} \sum_{ij} \gamma_{ij}\mathcal{D}(q(\boldsymbol{z^s}|\boldsymbol{x^{s}});q(\boldsymbol{z^t}|\boldsymbol{x^{t}})) + \beta\mathcal{D}_{KL}^{t}(q(\boldsymbol{z}^{t}|\boldsymbol{x}^{t})||p_{\theta}(\boldsymbol{z^{t}})), \label{eq4}
\\
\min_{\gamma \in \prod, q(\boldsymbol{z}|\boldsymbol{x})
} \sum_{ij} \gamma_{ij}\mathcal{D}(q(\boldsymbol{z^s}|\boldsymbol{x^{s}});q(\boldsymbol{z}^{t}|\boldsymbol{x}^{t})) + \beta\mathcal{D}_{KL}^{t}(q(\boldsymbol{z}^{t}|\boldsymbol{x}^{t})||p_{\theta}(\boldsymbol{z}^{t})) + \mathcal{L}_{vae}^{s} \label{eq5}
\end{align}

\noindent where $\mathcal{D}(q(\boldsymbol{z^s}|\boldsymbol{x^{s}});q(\boldsymbol{z^t}|\boldsymbol{x^{t}})) = \alpha||q(\boldsymbol{z^s}|\boldsymbol{x^{s}}) -  q(\boldsymbol{z^t}|\boldsymbol{x^{t}})||^{2}_{2}$ is the squared Euclidean distance and  $\beta\mathcal{D}_{KL}^{t}$ regularizes the target distribution. The final objective of OLVA is formulated in Eq.~(\ref{eq5}), which optimizes jointly for: i) an embedding function $q(\boldsymbol{z}|\boldsymbol{x})$ that maps both the source and the target domain to a semantic latent space $\mathcal{Z}$ regularized to follow normal distribution; ii) a transportation matrix $\gamma$ that aligns similar semantic vectors $\boldsymbol{z}$ from both domains in the latent space; and iii) a predictive function $p(\boldsymbol{y}|\boldsymbol{z})$ for masks predictions.


\subsection{Learning OLVA}

With the formulation presented in in Eq.~(\ref{eq5}), our framework learns a common latent space that conveys aligned information for both the source and target domain. To solve Eq.~(\ref{eq5}), we use an alternating method ~\cite{OTCourty}. Therefore, we optimize $\gamma$, with fixed $q(\boldsymbol{z|x})$ and $p(\boldsymbol{y|z})$, which reduces to the problem in Eq.~(\ref{eq5}) to solving a classic OT problem with cost matrix $C_{i,j} = \alpha||q(\boldsymbol{z}^{s}|\boldsymbol{x}^{s})-q(\boldsymbol{z}^{t}|\boldsymbol{x}^{t})||^{2}_{2}$. Then, we optimize $q(\boldsymbol{z|x})$ and $p(\boldsymbol{y|z})$, with fixed $\gamma$, this turns the problem in Eq.~(\ref{eq5}) to a standard deep learning problem. Similar to Damodoran\etal~\cite{damodaran2018deepjdot}, we solve the optimization problem with a stochastic approximation using mini-batches of size $m + n$ from the source and target domains respectively, which leads us to the optimization problem presented in Eq.~(\ref{eq6}). The stochastic approximation yields a computationally feasible solution for both the OT and VAE. The discrepancy measure and the KL-Divergence regularization are computed at the latent space layer, while the segmentation loss uses the output layer.

\begin{dmath}
\min_{q,p}\mathbb{E}\left[ \frac{1}{m}\sum_{i=1}^{m}\mathcal{L}_{dice}^{s}(\boldsymbol{y}^{s},p(\boldsymbol{y}^{s}|\boldsymbol{z}^{s})) + \frac{1}{m}\sum_{i=1}^{m}\beta\mathcal{D}_{KL}^{s} + \frac{1}{n}\sum_{i=1}^{n}\beta\mathcal{D}_{KL}^{t} + \frac{1}{m}\sum_{i=1}^{m}||\boldsymbol{y}^{s} - p(\boldsymbol{y}^{s}|\boldsymbol{z}^{s})||^{2} + \min_{\gamma \in \Gamma(\zeta^{s},\zeta^{t})} \sum_{i,j}^{m+n} \gamma_{i,j} \alpha|| q(\boldsymbol{z}^{s}|\boldsymbol{x}^{s}) - q(\boldsymbol{z}^{t}|\boldsymbol{x}^{t})||^{2}
\right]
\label{eq6}
\end{dmath}


\textbf{Architecture and implementation details:} OLVA accept batches containing $128$ source and $128$ target samples. The input dimension is $256\times256\times3$. The encoder is composed of five convolutional layers, with stride by 2 for down-sampling, and with a leaky rectified linear unit (lrelu) activation, with a leakage rate of $0.3$. The number of feature maps is successively $32,32,64,64,\text{and}~64$. The last convolutional is flattened and mapped using a linear fully connected layer into two vectors $(\mu,\sigma)$, each composed of $K=128$ features followed by a dropout of rate $0.3$. A latent vector $\boldsymbol{z}$ is generate as $\mu + \sigma \odot \boldsymbol{\epsilon}$, where $\boldsymbol{\epsilon}\sim\mathcal{N}(0,I)$ and given as an input to the decoder. The decoder is composed of five up-convolutional layers, with a lrelu activation, each composed of $64,64,32,32,\text{and}~4$ feature maps. The output layer with a sigmoid activation provides a mask of shape $256\times256\times4$. A learning rate of $0.0001$ is used with Adam optimizer. Using the validation set we experimentally tuned, $\alpha=10$ to focus more on the alignment loss and $\beta=0.1$. The total number of iterations is 10,000. 


\section{Experiments and Results}


We use the public MM-WHS dataset \cite{zhuang2016multi} for cardiac segmentation consisting of $20$ MR ($\sim128$ slices) and $20$ CT ($\sim256$ slices) unpaired and multi-site images from $40$ patients. We followed the state-of-the-art data processing, domain adaptation protocol and evaluation metrics \cite{huo2018synseg,ouyang2019data,dou2018pnp,chen2019synergistic,chen2020unsupervised}. For data processing, we use the coronal view slices, cropped to $256 \times 256$ and normalized to zero mean and unit variance. To consider contextual information three adjacent slices ($256 \times 256 \times 3$) were stacked at the input and the middle slice label was used as the ground truth. Data augmentation included rotation, scaling, and affine transformations. A total of $11998$ MR and $9598$ sub-volumes were generated (each $256\times256\times3$). For domain adaptation, we randomly split each modality into training (16 subject) and testing (4 subjects). We use MR as a source domain, with $9599$ sub-volumes for training and $2399$ for validation. We set CT as a the target domain, with $8399$ sub-volumes for training and $1199$ for evaluation. We report the performance in terms of Dice Similarity Coefficient (DSC) and the Average Symmetric Surface Distance (ASSD).

\begin{figure}[b]
\includegraphics[width=0.9\textwidth]{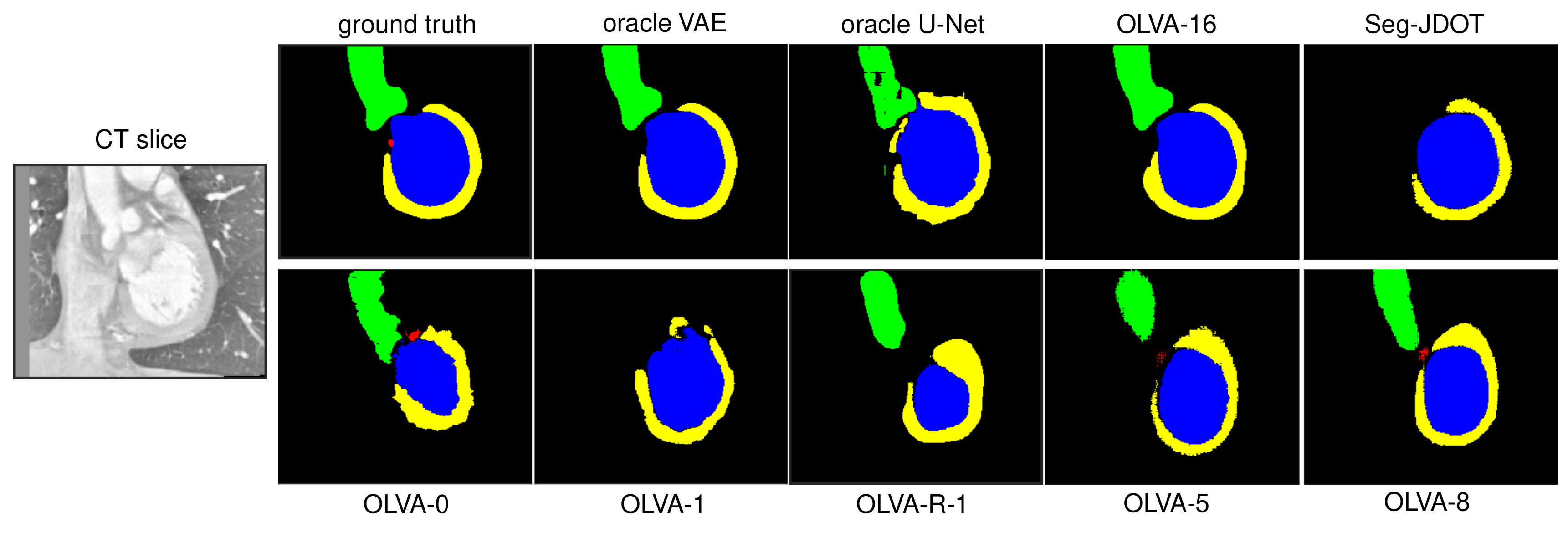}
\caption{Qualitative results of adaptation performances on segmentation.} \label{fig3}
\end{figure}

\textbf{Experimental Settings and Results:} We consider four experimental settings. \emph{First}, we compare our model without optimal transport and trained with full supervision over the CT images (oracle VAE). We compare this setting against a U-Net~\cite{ronneberger2015u} to show how our VAE constrains the shape of the predictions to be valid. The oracle VAE also serves as an upper-bound baseline. \emph{Second}, to illustrate the domain shift problem, we consider the situation when no adaptation is performed, thereby, we train VAE-0 and U-Net-0 over MR images and evaluated them over CT images. \emph{Third}, we consider the SOA setting, in which 16 labeled and 16 unlabeled source and target sequences are used (OLVA-16). We compare this setting with four SOA methods for medical UDA: PnP-AdaNet \cite{dou2018pnp}, SIFA \cite{chen2019synergistic}, Synseg-net \cite{huo2018synseg} and Seg-DJOT \cite{ackaouy2020unsupervised}. We also compare to two SOA methods for natural image UDA: CycleGAN \cite{isola2017image} and AdaOutput \cite{tsai2018learning}. \emph{Fourth}, we consider a more ambitious scenario where we assume that only one unlabeled target sequence is available (OLVA-1). Therefore, we randomly draw one scan from the target set. To train OLVA-1 and to avoid overfitting, we fix all the model parameters and only the fully connected layer is retrained using loss evaluated at the latent space of Eq.~\ref{eq6}. We also perform an experiment where an auxiliary reconstruction task~\cite{duque2020spatio} is integrated (OLVA-R-1). The quantitative and qualitative results are presented in Table~\ref{TAB1} and Fig.~\ref{fig3}

\begin{table}
\caption{Performances of UDA from MR to CT images under different settings. Postfix -0, -1 or -16 after names of each method indicate the number of unlabelled target scans used for training. We mark in bold the best results and we underline the second best. }
\small \setlength{\tabcolsep}{1.85pt}
\resizebox{0.85\linewidth}{!}{
\begin{tabular}{l *{5}{l} p{0.5\tabcolsep} *{5}{l}}
\toprule
 & \multicolumn{5}{c}{\textbf{DSC Score}} &&  \multicolumn{5}{c}{\textbf{ASSD Score (mm)}} \\
 \cmidrule(r){2-6} \cmidrule(l){8-12}
 \textbf{Methods} & LV-M & LA-B & LV-B & A-A & \textbf{avg} & & LV-M & LA-B & LV-B & A-A & \textbf{avg} \\

 \midrule
 oracle U-Net & 0.83 & 0.89 &  0.92 &  0.93 &  0.89 && 0.38 & 0.39 &  0.28 &  0.31 &  0.34 \\

oracle VAE & \textbf{0.95} & \textbf{0.97} &  \textbf{0.97} & \textbf{ 0.96} &  \textbf{0.96} && \textbf{0.06} & \textbf{0.04} &  \textbf{0.03 }&  \textbf{0.05 }&  \textbf{0.05} \\

  \midrule
 U-Net-0 & 0.10 & 0.27 &  0.02 &  0.24 &  0.15 && 36.0 & 19.4&  48.6 & 31.9 &  26.2\\  
VAE-0 & \textbf{0.41} & \textbf{0.51} &  \textbf{0.60} & \textbf{ 0.48} &  \textbf{0.49} && \textbf{2.51} & \textbf{2.21}& \textbf{ 2.44 }& \textbf{2.95} &  \textbf{2.53}\\   

 
  \midrule
 
   \textbf{OLVA-16} & \textbf{0.79} & \textbf{0.87} &  \textbf{0.88} &  \textbf{0.88} &  \textbf{0.85} && \textbf{0.56} & \textbf{0.53} &  \textbf{0.37} &  \textbf{0.45} & \textbf{0.31} \\

   Seg-DJOT-16 & 0.57 & 0.60 &  0.57 &  0.62 &  0.59 && 3.64 & 3.62 &  3.85 &  5.20 &  4.07 \\     
     SIFA-16 & \underline{0.58} & 0.76 &  \underline{0.76} &  \underline{0.81} &  \underline{0.73}  &&  \underline{3.44} & 3.83 & 3.30 &  2.64 &  {3.32}\\
    Pnp-AdaNet-16 & 0.50 & \underline{0.77} &  0.60 &  0.79 &  0.66 && 10.2 & 4.04 &  8.60 &  \underline{2.28} &  6.22 \\
    
     SynSeg-Net-16 & 0.41 & 0.69 &  0.52 &  0.72 & 0.58 && 4.60 & 3.80 &  3.40 &  5.60 &  4.35 \\   
    AdaOutput-16 & 0.43 & 0.76 &  0.54 & 0.65 & 0.59 && 4.68 &  \underline{2.89} &  \underline{3.10} &  6.15 &  4.20\\  
   CycleGAN-16 & 0.28 & 0.75 &  0.52 &  0.73 &  0.57 && 4.85 & 6.20 &  3.92 &  5.54 &  5.30 \\
  \midrule

   OLVA-1 & 0.58 & 0.69 &  0.64 &  \underline{0.67} &  0.64 && 2.10 & 1.95 &  1.85 & \textbf{2.30} &   \underline{2.05} \\    
   OLVA-R-1 & \textbf{0.68} & \underline{0.70} &  \underline{0.78} &  0.60 &  \underline{0.69} && \textbf{1.89} & \textbf{1.88} &  \textbf{1.51 }&  \underline{2.43} & \textbf{1.92}\\   
   
   DECM-1 & \underline{0.60} & \textbf{0.78} &  0.71 & \textbf{0.78 }&  \textbf{0.72} && 7.37 & 3.87& 6.44 &2.77&  5.11\\

    Seg-DJOT-1 & 0.19 & 0.25 &  0.21 &  0.20 &  0.21 && 9.64 & 13.7 &  8.18 &  10.3 &  10.4 \\    
    SIFA-1 & 0.39 & 0.53 &  \textbf{0.80} &  0.62 &  0.62 && 12.8 & 4.12 &  7.70 &  2.72 &  6.84 \\  
    Pnp-AdaNet-1  & 0.29 & 0.48 &  0.33 &  0.58 &  0.25 && 25.1 & 27.1 &  27.7 &  7.14 &  21.8\\ 
        


\bottomrule
\end{tabular} }\label{TAB1}
\end{table}

\textbf{Discussion:} Table~\ref{TAB1} shows that our supervised baseline method outperforms the U-Net, achieving a high DSC and more importantly producing valid cardiac shape predictions as seen in Fig.~\ref{fig3}, and as reflected by the ASSD score. When no adaptation is considered, VAE-0 achieves $49\%$ DSC, while U-Net achieved only $15\%$. As our VAE-0 pushes the latent semantic features to be close to normal distribution, it partially aligns the marginal distributions. In the UDA setting, OLVA-16 outperforms the SOA's best results by an additional $12.5\%$ in DSC! and having minimal erroneous prediction as seen in Fig.~\ref{fig3}, with average ASSD of $0.31~\text{mm}$. Considering the target data scarcity UDA Setting, OLVA-1 achieved the second best results after DECM-1 with an $8\%$ DSC difference. The results of OLVA-1 when the reconstruction auxiliary task is introduced (OLVA-R-1) reduce the gap to $5\%$ at the price of increasing the model complexity. The second-place is honorable, comparing the 1.7 million parameters of OLVA-1 with the more than 95 million parameters of DECM-1, and considering the quality of predictions as reflected by the ASSD. We also examine OLVA's performance when trained with randomly sampled 5 and 8 targets sequences. OLVA-5 achieves similar performance to DCEM-1, while OLVA-8 achieves better DSC score $79\%$. As for further ablation studies, we change the latent dimension to $64$, $256$ and $512$. With $K=64$, a degradation in the source domain performance was observed, yielding an average DSC score of $79\%$. With $K=256$, similar performances to $K=128$ is achieved. When $K=512$, a degradation in the performance over the source and the target domain is observed, leading to $69.6\%$ target DSC score for OLVA-16. This degradation is expected as optimizing for $\gamma$ requires a reasonable number of samples which grows with $K$'s dimensionality \cite{ackaouy2020unsupervised}. 


\section{Conclusion}
To improve the applicability of deep learning model on new modality where it is expensive to acquire expert annotations, unsupervised domain adaptation represents a central solution. In this paper, we tackle the problem of unsupervised cross-modality medical image segmentation 
with
a novel framework that jointly integrates VAE and OT theory to solve UDA problem. OLVA is a simple, efficient and lightweight model, which makes it practical to deploy in real-life without requiring a machine with huge computational resources. The usability of our method can be integrated within other learning regimes, for instance, a weakly-supervised model where sparse annotation of biomedical volumetric data are available and the aim would be to leverage the rest of the unlabeled data by matching them with the available labeled set. Future work will address the problem of building a general segmenter where the adaptation from one task to another is done with minimal task-specific information and to leverage other tasks information.




\bibliographystyle{splncs04}


\end{document}

%% file: bold.tex
\usepackage{amssymb}